\pdfoutput=1

\documentclass[11pt]{article}

\usepackage[]{acl}

\usepackage{times}
\usepackage{latexsym}
\usepackage{graphicx}

\usepackage[T1]{fontenc}

\usepackage[utf8]{inputenc}

\usepackage{microtype}

\usepackage{inconsolata}
\usepackage{booktabs}
\usepackage{pifont}
\newcommand{\cmark}{\ding{51}}%
\newcommand{\xmark}{\ding{55}}%

\title{Can GPT Redefine Medical Understanding? \\Evaluating GPT on Biomedical Machine Reading Comprehension}

\author{Shubham Vatsal, Ayush Singh \\
  inQbator AI at eviCore Healthcare \\
  Evernorth Health Services \\
  \texttt{firstname.lastname@evicore.com}}
  
\begin{document}

\maketitle

\begin{abstract}
Large language models (LLMs) have shown remarkable performance on many tasks in different domains. However, their performance in contextual biomedical machine reading comprehension (MRC) has not been evaluated in depth. In this work, we evaluate GPT on four contextual biomedical MRC benchmarks. We experiment with different conventional prompting techniques as well as introduce our own novel prompting method. To solve some of the retrieval problems inherent to LLMs, we propose a prompting strategy named Implicit Retrieval Augmented Generation (RAG) that alleviates the need for using vector databases to retrieve important chunks in traditional RAG setups. Moreover, we report qualitative assessments on the natural language generation outputs from our approach. The results show that our new prompting technique is able to get the best performance in two out of four datasets and ranks second in rest of them. Experiments show that modern-day LLMs like GPT even in a zero-shot setting can outperform supervised models, leading to new state-of-the-art (SoTA) results on two of the benchmarks. 
\end{abstract}


\begin{table*}[h]
    \centering
    \begin{tabular}{lcccc}
    \toprule
        \textbf{Dataset}     & \textbf{ProcessBank} & \textbf{BioMRC}  & \textbf{MASH-QA} & \textbf{CliCR}   \\
        \midrule
        \# QA Pairs & 150     & 6250     & 3493 & 7184\\
        Avg Context Length    & 85     & 255   & 863  & 1461 \\
        Max Context Length    & 266    & 510  & 2911 & 3952\\
    \end{tabular}
    \caption{Corpus Level Statistics}
    \label{tab:datset}
\end{table*}

\section{Introduction}
Machine Reading Comprehension (MRC) is defined as a task where a system tries to answer a question based on a given context. The context could be anything ranging from a couple of passages to a list of documents. Even though much research has been conducted on MRC, several challenges remain when dealing with MRC tasks \cite{sugawara2022makes}, such as the inability to handle long-range dependencies when trying to do reasoning and domain adaptation. Recent improvements in large language modeling has alleviated a lot of the aforementioned issues.

MRC in the biomedical domain \cite{hermann2015teaching, baradaran2022survey} has always been a key area of research.
Solving a biomedical MRC task faces various challenges including large intricate in-domain vocabulary, dependency on global knowledge, etc.
Due to these challenges, there is a wide gap between the performance of conventional methods in the general domain and that of the biomedical domain. Although, traditional machine learning models did show some improvement, they have never been any close to human baselines or gold standards. Contrary to this preconceived notion, modern-day LLMs have shown remarkable performance on many biomedical tasks \cite{nori2023capabilities, yang2023enhancing, cheng2023artificial}.


MRC can have different variations in itself. A contextual MRC requires the LLMs to answer a query solely by relying on a given context.
In contrast, a context-free MRC relies on model's embedded knowledge base or any open-source knowledge base, such as Wikipedia, to answer a query instead of using only the context provided. Some of the datasets corresponding to  context-free MRC are \citet{zhang2018medical,pal2022medmcqa}. These datasets are classified under context-free MRC because the given context is not sufficient to answer the questions. There is a definite need to explore the LLM's inherent knowledge base or any other source of knowledge to answer these queries. Similarly, \citet{berant2014modeling,pappas2018bioread,zhu2020question} comprise of the datasets for contextual MRC. Again, these datasets have been categorized under contextual MRC because the queries asked can be correctly answered just by looking at the provided context. There is no need to induce any kind of external knowledge in order to answer these questions.

Recent LLMs have attained unprecedented performance in a wide array of natural language processing (NLP) tasks \cite{chang2023survey,vatsal2024survey}. Although their performance have been evaluated on a multitude of MRC benchmarks in a context-free setting, their performance in a contextual setting has been understudied. In this work, we fill out this missing gap by evaluating GPT \cite{openai2023gpt4} on standard contextual MRC benchmarks of the biomedical domain. The key contributions are:


\textbf{1.} We evaluate different prompting techniques with GPT on four contextual biomedical MRC benchmarks and report new SoTA results.

\textbf{2.} We propose a novel prompting method \textit{Implicit RAG}. In this method, the LLM is asked to first retrieve the sections or textual extracts from the context that may be relevant to the query and then answer the given query. This technique shows that unlike conventional RAG we no longer need vector databases to store the embeddings of the entire corpus. It further emphasizes that LLMs are capable enough to do the retrieval in one go. Experiments show that this technique is able to achieve the best results in two out of four discussed datasets and ranks second in rest of them.

\textbf{3.} Although machine evaluation is a good measure of performance, it falls short when evaluating artificially generated text \cite{schluter-2017-limits}, where actual human preferences are significantly superior. Therefore, we report qualitative preference metrics by human experts on the output of our proposed approach \textit{Implicit RAG}. We find that humans agree with the generated outputs most of the time.

\begin{figure*}
\fbox{\begin{minipage}{41em}
You are a \textit{\{profession\}} who is given a \textit{\{context\_type\}} and a corresponding \textit{\{query\_type\}}. Your job is to read the given \textit{\{context\_type\}} and then select the best option from a list of options to answer the \textit{\{query\_type\}}.

\hfill\break
The \textit{\{query\_type\}} that needs to be answered is listed below.

\hfill\break
\textit{\{query\_type\}}: \textit{\{query\_text\}}
\hfill\break
List of options: \textit{\{options\}}

\hfill\break
Here is the \textit{\{context\_type\}} that needs to be read to select the best option from a given list of options for the \textit{\{query\_type\}}.

\hfill\break
\#\#\#
\textit{\{context\_text\}}
\#\#\#

\end{minipage}}
\caption{Basic Prompt Template}
\label{fig:basic}
\end{figure*}

\section{Related Work}
MRC evaluates a system's ability to comprehend and then reason to answer questions over the natural language present in a passage or context. Over the years, quite a few variations of this task have been devised to address and evaluate various aspects of a MRC system namely
cloze-style \cite{hermann2015teaching, yagcioglu2018recipeqa, pappas2018bioread, pappas2020biomrc}, multiple-choice \cite{richardson-etal-2013-mctest, lai2017race, Berant2014ModelingBP}, extractive \cite{yang2015wikiqa, trischler2016newsqa, zhu2020question} and generative QA \cite{nguyen2016ms, kovcisky2018narrativeqa}.
In this study, we strive to evaluate three of the above discussed forms of MRC in the biomedical domain which are cloze-style, extractive and multiple-choice using GPT.

In order to elicit an answer from an LLM like GPT, one needs to prompt it in natural language in an optimal manner to retrieve the intended answer. To that end, there has been tremendous development in finding optimal methods for prompting LLMs. The maximum performance boost has been seen from the Chain-of-Thought (CoT) Reasoning \cite{wei_chain-thought_2022} prompting strategy which asks the LLM to explain how it arrived at the answer. 
More recently, Analogical Reasoning (AR) \cite{yasunaga_large_2023} has been proposed that achieves drastically better performance than CoT and other prompting techniques. AR works by asking the LLM to reason about a problem by giving analogies which in return forces the model to leverage the global knowledge encoded in it. While prompting methods like CoT and AR improve LLM's performance by exploiting the model's global knowledge embedded in it, 
there has been an increase in developing novel techniques, especially for cases where the context that needs to be searched through to answer the asked query is huge. The context could be one huge document or a combination of multiple short/long documents. In such scenarios, it is very important to identify only the relevant chunks of the context required for the underlying task and pay attention to them. These emerging methods come under the umbrella of Retrieval Augmented Generation (RAG) \cite{lewis2020retrieval}, which has been shown to improve the performance of LLMs by retrieving contextually relevant information from corpora. The basic methodology behind RAG is to use, embed, and store context in a vector database. These embeddings can then be retrieved based on their semantic similarity to the query. 

All the aforementioned prompting methods help interface and contextualize inputs in a better manner for LLMs. While the efficacy of these methods has been seen on several large benchmarks in different domains, however, the degree to which they help in contextual biomedical MRC has been understudied. \citet{mahbub2022bioadapt} presented an adversarial learning-based domain adaptation framework for the biomedical MRC task to address the discrepancies in the marginal distributions between the datasets of the general and biomedical domains.  \citet{nori2023capabilities} evaluated GPT on medical competency examinations and benchmark datasets. Even though their work talks about biomedical MRC, it concentrates only on context-free MRC benchmarks. Similarly, \citet{singhal2023towards} evaluated Med-PaLM 2 on medical competency examinations and thus focuses on context-free biomedical MRC. 



\section{Datasets}
The four datasets from the biomedical and healthcare domain we choose to explore and analyze the performance of GPT are ProcessBank \cite{berant2014modeling}, BioMRC \cite{pappas2020biomrc}, MASH-QA \cite{zhu2020question} and CliCR \cite{vsuster2018clicr}. There are a multiple reasons for selecting these four datasets. First, we want to focus on datasets that have not yet been evaluated by modern-day LLMs like GPT. Next, we want to pick up datasets that vary in their statistics and nature. Finally, based on our understanding, these 4 datasets covered the majority of research around contextual MRC in the biomedical field.

ProcessBank contains descriptions of biological processes as context accompanied by multiple-choice questions. BioMRC, an improved version of BioREAD \cite{pappas2018bioread} is a large-scale cloze-style dataset. It contains abstracts and titles of biomedical articles and the task of any MRC system is to predict the missing entity in a title using the corresponding abstract as context. In BioMRC, all the biomedical entities mentioned in the abstract are considered as candidate answers and thus one option needs to be chosen from them. MASH-QA is associated with consumer health domain where the answers can consist of sentences from multiple spans of the long context. The candidate answers include every single sentence of the given context. CliCR is again a cloze-style dataset. It contains cloze queries from clinical case reports. Unlike BioMRC, CliCR doesn't really have a list of candidate answers with one of them being the correct answer. Rather, CliCR contains a ground-truth answer set which consists of different lexical and semantic variations of the ground-truth answer, and thus all of them are correct. We use only the test sets of these datasets for all our prompting experiments in a zero-shot setting. We use BioMRC LITE version of BioMRC. The statistics of the four datasets are listed in Table \ref{tab:datset}.
\begin{figure*}[h]
    \includegraphics[scale=0.58]{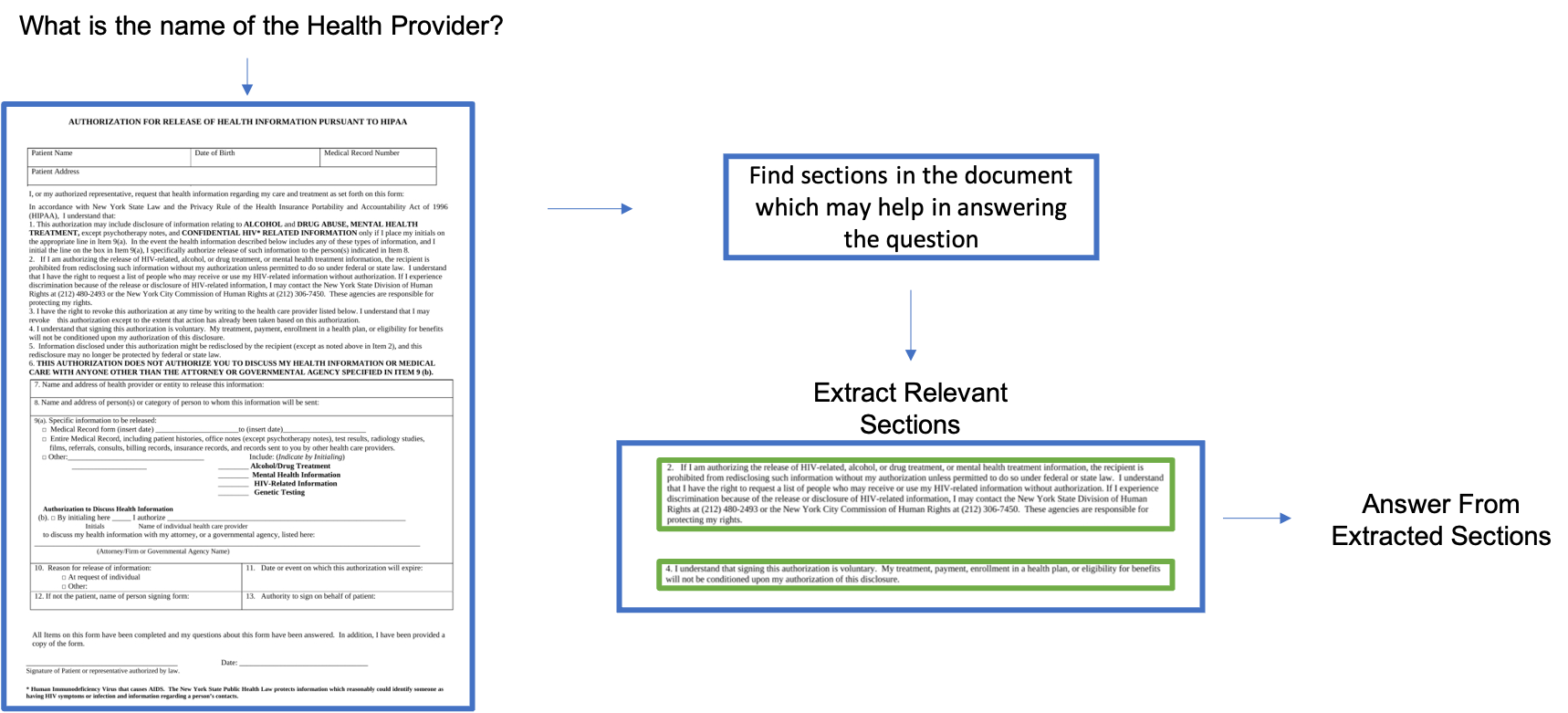}
    \caption{Implicit RAG Technique}
    \label{fig:iragtech}
\end{figure*}

\begin{figure*}
\fbox{\begin{minipage}{41em}
You are a \textit{\{profession\}} who is given a \textit{\{context\_type\}} and a corresponding \textit{\{query\_type\}}. Your job is to read the given \textit{\{context\_type\}} and then select the best option from a list of options to answer the \textit{\{query\_type\}}.

\hfill\break
The \textit{\{query\_type\}} that needs to be answered is listed below.

\hfill\break
\textit{\{query\_type\}}: \textit{\{query\_text\}}
\hfill\break
List of options: {\textit{options}}

\hfill\break
Identify \textit{\{number\_of\_sections\}} most relevant sections or text extracts from the given \textit{\{context\_type\}} that may help in selecting the best option to answer the given \textit{\{query\_type\}}. The identified sections or text extracts should be distinct from each other. The identified sections or text extracts must be between \textit{\{lower\_limit\_length\}} to \textit{\{upper\_limit\_length\}} words long.

\hfill\break
Now, choose the best option to answer the given \textit{\{query\_type\}} using the identified sections or text extracts.  

\hfill\break
Here is the \textit{\{context\_type\}} that needs to be read to select the best option from a given list of options for the \textit{\{query\_type\}}.

\hfill\break
\#\#\#
\textit{\{context\_text\}}
\#\#\#
\end{minipage}}
\caption{Implicit RAG Prompt Template}
\label{fig:irag}
\end{figure*}







\section{Prompting Techniques}
 While an exhaustive study of all prior prompting strategies could have been a better experimental setup but due to the cost-prohibitive nature of running large-scale experiments on GPT, we only select the techniques that have shown to perform well in the general domain. Along with these strategies, we also introduce a novel prompting method named \textit{Implicit RAG}. We elaborate on all of these different prompts and their corresponding templates. There may be slight differences in prompt templates of the same prompting strategy across different datasets in order to adhere to the syntactic and semantic rules of English grammar as well as align with the dataset characteristics.

\paragraph{Basic}
\label{basic}

The prompt template used for this technique is shown in Figure \ref{fig:basic}. The Basic prompting approach asks GPT to answer the query in the simplest way possible. The \textit{profession} placeholder specifies the role that GPT has to take in order to answer the asked question. Based on the source of the dataset, this placeholder takes the value of \textit{biologist} in the case of ProcessBank, \textit{biomedical researcher} in the case of BioMRC, \textit{consumer healthcare expert} in case of MASH-QA  and \textit{medical expert} in the case of CliCR dataset. Again, based on the source of the dataset, the placeholder \textit{context\_type} takes the value of \textit{paragraph} for ProcessBank, \textit{abstract of the paper} for BioMRC, \textit{healthcare article} for MASH-QA and \textit{clinical case report} for CliCR. The placeholder \textit{query\_type} takes the value of \textit{query} for ProcessBank, \textit{title containing the missing entity} for BioMRC, \textit{query} for MASH-QA and \textit{query containing the missing entity} for CliCR. The \textit{query\_text} placeholder contains the actual text of the query and similarly \textit{context\_text} contains the actual text of the context. The \textit{options} placeholder is present only for ProcessBank and BioMRC datasets and contains the choices to select from while answering the asked query.


\paragraph{Chain-of-Thought Reasoning (CoT)}
The rationale behind using the CoT technique is that there may be multiple smaller questions that need to be answered first in order to conclude the answer of the final asked question. For example, one of the questions asked to GPT is \textit{Has there been at least 6 weeks of provider-directed conservative treatment?}. This question can easily be divided into 3 smaller questions \textit{Has there been any conservative treatment?}, \textit{Was the treatment provider-directed?} and \textit{What was the duration of conservative treatment?}. The prompt template used for this technique is exactly same to that of Figure \ref{fig:basic} with an additional line instructing the model to \textit{Think step by step}. 

\paragraph{Analogical Reasoning (AR)}

Inspired by \citet{yasunaga2023large}, we design our own analogical reasoning strategy by tweaking the prompt to fit our problem statement. We do this because, unlike the general domain, GPT would not be able to recall specific dataset-level knowledge, as we are not sure if it was ever trained on the datasets being used in our study. Rather, we frame the prompt so that GPT does not need to rely on a lot on global knowledge. To that end, instead of asking GPT to generate any kind of relevant QA pairs based on its global knowledge, we ask GPT to generate QA pairs from the given context and then answer the initial question. There is one hyperparameter for this technique which is the number of QA pairs to generate. 



\paragraph{Implicit Retrieval Augmented Generation (RAG)}
Most of the work on RAG talks about data retrieval based on an accepted relevancy score and then using LLM prompts to answer the given query. The data retrieval is done by storing the embeddings from some encoder of the entire corpus (a datapoint's context in our case) in a vector database index and then retrieving the most matching data points (text extracts or sections from a datapoint's context) for a given query. The key idea behind using RAG is that it helps in saving a lot of computational cost and improves the LLM's performance as now it has to look in a smaller knowledge space to answer the asked question. In our proposed novel prompting technique \textit{Implicit RAG}, we completely ignore the overhead involved in getting the embeddings of the entire corpus and storing them in a vector database. Instead, we ask the LLM itself to find the most relevant text extracts or sections in the given context which may help in answering the asked question, and then later use these extracted sections to conclude the answer to the original question. The general working of our proposed prompting technique is shown in Figure \ref{fig:iragtech}. There are two hyper-parameters for this technique. First is, the number of sections to extract, and the next one is the number of words in each section or text extract. The prompt template used for this technique is shown in Figure \ref{fig:irag}. The hyper-parameter values for the number of sections and number of words in each section is provided in the placeholders \textit{number\_of\_sections} and \textit{lower\_limit\_length} and \textit{upper\_limit\_length}.



\section{Results \& Analysis}
\footnote{We will release the relevant sections identified by the Implicit RAG technique as well as the question-answer pairs generated by the AR method upon acceptance for all the discussed datasets which can prove useful for other researchers.}We use the 32k context window version of GPT-4 to conduct all our experiments. 
We set the temperature, frequency penalty, and presence penalty to 0 and max tokens to 1000 for GPT-4. The results for all the datasets have been discussed individually below. Based on different iterations of experiments, we choose the hyper-parameter number of QA pairs to generate for AR to be 3 for all the datasets.
Similarly, for \textit{Implicit RAG}, we choose the hyper-parameter \textit{lower\_limit\_length} and \textit{upper\_limit\_length} values as 50 and 200 respectively for all the datasets except MASH-QA. For MASH-QA, we choose \textit{lower\_limit\_length} as 0 and \textit{upper\_limit\_length} as 300. We choose \textit{number\_of\_sections} for \textit{Implicit RAG} to be 1 for MASH-QA, 2 for ProcessBank and 3 for BioMRC and CliCR. 
\paragraph{ProcessBank}
\begin{table}[]
    \centering
    \begin{tabular}{@{}lc@{}}
    \toprule
    \textbf{Method}     & \textbf{Accuracy}   \\ \midrule
    Basic (Full)        & 0.96 \\
    CoT (Full)          & 0.96 \\
    AR (Full)           & 0.96  \\
    Implicit RAG (Full) & \textbf{0.97} \\ \midrule
    Implicit RAG (Full) & \textbf{0.97} \\
    Gold Structure  & \textbf{0.77} \\
    ProRead  & 0.67 \\
    SyntProx  & 0.60 \\
    TextProx  & 0.55 \\
    Bow       & 0.47 \\ 
    \end{tabular}
    \caption{Results on ProcessBank. The results for Gold Structure, ProRead, SyntProx, TextProx and Bow have been discussed in \citet{berant2014modeling}}
    \label{tab:processbank}
\end{table}

The results for ProcessBank are shown in Table \ref{tab:processbank}. We ran all 4 prompting strategies on the entire test set of 150 datapoints in a zero-shot setup. Every single prompting method outperforms the previously proposed methods, thus giving us a new SoTA on this dataset. Among the different prompting strategies, \textit{Implicit RAG} gets the best results. The important observations are: 

\textbf{1.} The only 4 to 5 datapoints that GPT got wrong either are very confusing for even humans to answer or had some typo or extra punctuation in ground-truth answers which GPT was not able to mimic during its generation.

\textbf{2.} It is observed that all the GPT prompting strategies work more or less the same if the question can be answered from a small span in the provided context. The reason why \textit{Implicit RAG} is able to outperform other techniques is because this dataset includes around 30\% of temporal and true-false type questions which require extensive analysis of the entire context and that the answer can be spread in different segments of the context. Therefore, reducing the knowledge space by extracting relevant sections to answer the asked question helps in improving the performance.


\paragraph{BioMRC}

The results for BioMRC are listed in Table \ref{tab:biomrc}. Due to cost-related reasons, we first compare different prompting methods by running them on a randomly selected 15\% (1000 datapoints) subset of the test set and then choosing the best prompting technique to run on the entire test set. All these experiments are done in a zero-shot setting. Amongst the different prompting techniques, Basic prompting gets the best results and \textit{Implicit RAG} ranks second. The important observations are:

\textbf{1.} Even though BioMRC is a cleaner version of BioREAD, there are still elements of lack of structure in the dataset. For example, there is no 1-1 mapping between entity IDs and entities. So, this means the same entity can be mapped to multiple entity IDs and vice versa which causes a lot of confusion when quantifying performance. The authors of BioMRC claim that for any query, the abstract or the context contains all the candidate options including the correct answer but this is not always true leading to more confusion during evaluation.


\textbf{2.} Quite a few times GPT is able to generate an acronym answer instead of its corresponding full form. Ideally, both acronyms and their full forms must be considered as correct answers.

\textbf{3.} There are instances where GPT being a generative model is able to produce semantically similar answers but still they are marked wrong as they do not exactly match with the correct answer. An embedding based metric can be helpful here.

\textbf{4.} There are a lot of entities which are semantically and syntactically the same but still belong to different ontologies and thus have different entity IDs. For example, in one case, GPT generates the answer \textit{amino acids} where the correct answer is \textit{amino acid} but since both of these entities have different IDs, this answer had to be marked wrong.

\begin{table}[]
    \centering
    \begin{tabular}{@{}lc@{}}
    \toprule
    \textbf{Method}       & \textbf{Accuracy}   \\ \midrule
    Basic (1000)        & \textbf{0.87} \\
    CoT (1000)            & 0.81 \\
    AR (1000)          & 0.82  \\
    Implicit RAG (1000) & 0.83 \\ \midrule
    Basic (Full)        & \textbf{0.87} \\
    MLP-based Weighting  & \textbf{0.88} \\
    AoA-Reader with BioBERT  & 0.87 \\
    SciBERT-Max-Reader  & 0.80 \\
    AoA-Reader  & 0.70 \\
    AS-Reader  & 0.62 \\ 
    \end{tabular}
    \caption{Results on BioMRC. The results for models MLP-based Weighting and AoA-Reader with BioBERT are discussed in \citet{lu2022contextual} whereas the results for SciBERT-Max-Reader, AoA-Reader and AS-Reader are explained in \citet{pappas2020biomrc}  }
    \label{tab:biomrc}
\end{table}

Another important aspect to note here apart from the lack of structure in the dataset is the overall system design of supervised models which are being compared to GPT when talking about SoTA. Supervised models use 70\%-80\% of the available data as their training set which allows their parameters to get a good idea about the nuances of the dataset whereas in case of GPT, all our experiments are being conducted in a zero-shot setting. Also since GPT is a generative model, the chances of GPT generating an answer not present in the candidate answer list despite the final answer being semantically and syntactically the same is really high. But a supervised model is never going to face this problem as it makes its prediction based on the confidence score for each candidate answer and thus the final answer is always going to be present in the candidate answer list.

\begin{table}[]
    \centering
    \begin{tabular}{rcccc}
    \toprule
    \textbf{Method}  & \textbf{EM} & \textbf{F1} & \textbf{P} & \textbf{R} \\ \midrule
    Basic (600)        & \textbf{0.12} & \textbf{0.53} & 0.50 & \textbf{0.56} \\
    Analogical (600)   & 0.11 & 0.50 & \textbf{0.53} & 0.47 \\
    CoT (600)          & 0.11 & 0.52 & 0.50 & 0.55 \\
    Implicit RAG (600) & 0.10  & 0.52 & 0.51 & 0.52 \\ \midrule
    Basic (Full)       & \textbf{0.14} & \textbf{0.53} & \textbf{0.50} & \textbf{0.57} \\
    Bert         & 0.09  & 0.25 & 0.56 & 0.16 \\ 
    RoBERTa       & 0.09 & 0.29 & 0.58 & 0.19 \\
    XLNet         & 0.09  & 0.29 & 0.56 & 0.20 \\ 
    MultiCo       & \textbf{0.22} & \textbf{0.57} & \textbf{0.58} & \textbf{0.56} \\
    Tanda        & 0.09 & 0.25 & 0.56 & 0.16 \\
    \end{tabular}
    \caption{Results on MASH-QA dataset. The results for Bert, RoBERTa, XLNet, MultiCo and Tanda have been talked about in \citet{zhu2020question}}
    \label{tab:mashqa-results}
\end{table}

\paragraph{MASH-QA}
The results for MASH-QA are shown in Table \ref{tab:mashqa-results}. We start by conducting a comparison between different prompting strategies by evaluating them on a randomly selected 15\% (600 datapoints) subset of the test set. These experiments are undertaken in a zero-shot setting. As we can see in Table \ref{tab:mashqa-results}, Basic prompting performs the best while \textit{Implicit RAG} ranks second. The important points to discuss here are:

\textbf{1.} The answers in QA pairs of MASH-QA are very subjective. The authors of this dataset have not specified any structured process that was followed by the healthcare experts when trying to answer the questions asked on a website from where this dataset was sourced in the first place. An in-depth analysis shows that even though GPT is able to extract better answers a lot of times, since it does not match with the ground truth answers, the evaluation metrics do not reflect it's true capabilities.

\begin{table}[]
    \centering
    \begin{tabular}{@{}lcc@{}}
    \toprule
    \textbf{Method}       &  \textbf{EM} &  \textbf{F1}   \\ \midrule
    Basic (1100)        & 0.37 & 0.53  \\
    Analogical (1100)   & \textbf{0.39} & \textbf{0.54} \\

    CoT (1100)          & 0.36 & 0.51  \\
    Implicit RAG (1100) & 0.38 & \textbf{0.54}  \\ \midrule
    Analogical (Full)   & \textbf{0.34} & \textbf{0.52}  \\
    Human Novice  & 0.31 & 0.45  \\
    Human Expert  & \textbf{0.35} & \textbf{0.54}  \\
    GA-Anonym    & 0.25 & 0.33\\
    GA-Ent       & 0.22 & 0.30 \\
    GA-NoEnt    & 0.15 & 0.34 \\
    SA-Anonym   & 0.20 & 0.27 \\
    Sim-Entity  & 0.21 & 0.29 \\
    
    \end{tabular}
    \caption{Results on CliCR dataset. The results for Human Novice, Human Expert, GA-Anonym, GA-Ent, GA-NoEnt, SA-Anonym and Sim-Entity are explained in \citet{vsuster2018clicr}  }
    \label{tab:clicr}
\end{table}

\textbf{2.} There is a correlation observed between increase in the number of sections and decreasing performance of \textit{Implicit RAG}. The reason is that the answers in this dataset are long span and thus with increasing number of sections, there is a loss of contextual continuity as the ground truth answers can get split across multiple sections. This ends up confusing the LLM making it difficult to choose the right set of sentences from different sections. Hence, it performs the best when asked to extract just one section from the context. But because MASH-QA contains answers which can be present in disjoint spans of the context, extracting just one section is not able to make \textit{Implicit RAG} the best performing prompting method.

\begin{table*}

\centering

\begin{tabular}{ccccccccc}
\toprule
\textbf{Dataset} &\multicolumn{2}{c}{ProcessBank (50)} &\multicolumn{2}{c}{BioMRC (50)} &\multicolumn{2}{c}{MASH-QA (50)} &\multicolumn{2}{c}{CliCR (50)} \\
\midrule
\textbf{Pattern} & \cmark (46) & \xmark (4) & \cmark (41) & \xmark (9) & \cmark (7) & \xmark (43) & \cmark (31) & \xmark (19) \\
\midrule

Right Section & 100\% & 100\%  & 95\% & 56\% & 100\% & 93\% & 81\% & 32\% \\
Wrong Section & 0\% & 0\%  & 5\% & 44\% & 0\% & 7\% & 19\% & 68\% \\


\bottomrule
\end{tabular}
\caption{Qualitative Analysis of \textit{Implicit RAG} on ProcessBank, BioMRC, MASH-QA and CliCR}
\label{compare}
\end{table*}

\textbf{3.} One question which may arise is whether extracting one section or having the hyper-parameter number of sections set to 1 for \textit{Implicit RAG} makes it same as that of Basic prompting. \textit{Implicit RAG} and Basic prompting become the same only when we are not only extracting just one section from the context but also when the hyper-parameter number of words for \textit{Implicit RAG} is set equal to the length of the entire given context. But for MASH-QA, the hyper-parameter number of words is set to 300 with the lower limit being 0 and upper limit being 300 and hence they are different.

Again, we need to reiterate that GPT's performance in case of MASH-QA is being compared to supervised models which use 70\%-80\% of the total data as their training set allowing their parameters to capture granular details better than a generic LLM like GPT in a zero-shot setting.

\paragraph{CliCR}
The results for CliCR can be seen in Table \ref{tab:clicr}. Again, due to cost-related reasons, we first compare different prompting techniques by running them on randomly selected 15\% (1100 datapoints) subset of the test set and then choosing the best prompting method to run on the entire test set. All these experiments are done in a zero-shot setting. Amongst the different prompting strategies, \textit{Implicit RAG} and AR get the best results in terms of F1 metric. AR minutely performs better than \textit{Implicit RAG} when compared in terms of the Exact Match (EM) metric. However, EM is a very harsh metric for a generative model as there can be so many possible variations of semantically similar output which are not wrong. Since AR is computationally faster with respect to \textit{Implicit RAG}, we ran AR on the entire dataset. All the prompting methods outperform the previously proposed methods. Not only does GPT surpass the performance of previous models, but it also comes close to Human Expert performance while beating Human Novice results. 
The important observations are:

\textbf{1.} The authors of CliCR mention that for the training of supervised models, only those instances are used for which at least one ground-truth answer from the set of ground-truth answers occurs in the clinical case report or the context. But for the evaluation part, for both validation and test sets even those datapoints are included where there is no intersection between ground-truth set and the entities mentioned in the context. This favors supervised learning settings as supervised models have a separate training and development set which can allow the parameters of the model to learn such cues. GPT is still able to perform better possibly because the global knowledge embedded in its parameters gives it enough evidence to perform well.

\textbf{2.} The authors of CliCR compare various skills in their work between the previous SoTA (GPT is the new SoTA) model GA-NoEnt and Human Expert and show that there still exists a huge gap between them. Since GPT is able to achieve almost Human Expert level performance, we can expect it to show similar capabilities in other MRC tasks.

\textbf{3.} There are multiple reasons why \textit{Implicit RAG} performs well on this dataset. First, the mean length of context in this dataset is 1461 words which indicates that with increasing size of context, the chances of analysis of different sections of the context simultaneously to answer a question is high and that is the core idea behind \textit{Implicit RAG}. Next, the authors of CliCR list out that 70\% of the queries in this dataset require the \textit{bridging} skill, 40\% require the skill of \textit{tracking} and around 25\% demand the \textit{spatiotemporal} skill. All these three skills indicate that answering queries in this dataset require deriving cues from different segments of the context and that is what we propose as the key rationale behind \textit{Implicit RAG}.

\paragraph{Implicit RAG}
Out of the four datasets that we use in this study, \textit{Implicit RAG} is able to achieve the best results for two of them when compared with other prompting techniques. It ranks at the second place for the other two datasets. One of the questions which may arise is whether Implicit RAG can be applicable to contexts which cannot fit in LLM's 32k token limit. Implicit RAG will especially perform better than other prompting techniques in cases where the context size is more than 32k. In such cases, we can chunk the context and make multiple calls to Implicit RAG to retrieve relevant sections given the query. Once all the relevant sections have been retrieved, the last call to Implicit RAG can use these sections to arrive to an answer. But all other prompting techniques require analysis of the entire context (greater than 32k in this case) at the same time to arrive to an answer. We further do a qualitative analysis on 50 randomly picked datapoints for all four datasets. We check how many times the extracted sections are relevant to the question or not. Even if 1 out of all the extracted sections are relevant, we consider that to be a valid retrieval irrespective of whether the final answer was correct or incorrect. The results are shown in Table \ref{compare}. As we can see, \textit{Implicit RAG} is able to extract relevant sections in most cases.






\section{Conclusion}
In this work, we show that even in a zero-shot setting, GPT surpasses the performance of supervised models for two out of four benchmarks. Furthermore, GPT's performance comes close to that of Human Expert for one of the benchmarks. Our study corroborates that LLMs indeed have surpassed preconceived techniques even on difficult to model domains like biomedicine. We also come up with a novel prompting method \textit{Implicit RAG} which 
gets the best results in two out of four datasets and ends up at rank two in others. This opens a new research direction for the RAG domain allowing other researchers to experiment with this technique on other domain datasets.


\section{Limitations}
Due to cost associated with running large-scale experiments with GPT, we did a comparison of different prompting techniques on a subset of about 15\% of the entire test set for three out of four datasets we discuss in this work. It could be possible that there may be a slight difference in the distribution of the random subset we chose in comparison to the entire test set and this could potentially change the final results obtained by a given prompting technique although we expect the difference to be small. As discussed earlier, in cases where the answer to a query can be found in a small span of the context, there is not a huge difference between different prompting techniques. Thus, running the Basic prompting method will be computationally more inexpensive than running heavier prompting strategies like AR or \textit{Implicit RAG}. 

\bibliography{acl}



\end{document}